\documentclass[journal, twocolumn]{IEEEtran}
\usepackage[margin=1.0in]{geometry}  % change page margins

\usepackage{microtype}
\usepackage{graphicx}
\usepackage{subfigure}
\usepackage{amsfonts, amssymb, amsmath, amsthm}
\usepackage{picins}
\usepackage[ruled, vlined]{algorithm2e}
\usepackage{url}

\newcommand{\bsym}{\boldsymbol} % Copy definition of \boldsymbol into \bsym

\title{Variation-Incentive Loss Re-weighting for Regression Analysis on Biased Data}

\author{Wentai~Wu, Ligang~He, and Weiwei~Lin
\thanks{W. Wu, L. He (corresponding author, ligang.he@warwick.ac.uk) are with the Department of Computer Science, University of Warwick. W. Lin is with the School of Computer Science and Engineering, South China University of Technology.}
}% <-this % stops a space

\begin{document}

\maketitle

\begin{abstract}
Both classification and regression tasks are susceptible to the biased distribution of training data. However, existing approaches are focused on the class-imbalanced learning and cannot be applied to the problems of numerical regression where the learning targets are continuous values rather than discrete labels. In this paper, we aim to improve the accuracy of the regression analysis by addressing the data skewness/bias during model training. We first introduce two metrics, uniqueness and abnormality, to reflect the localized data distribution from the perspectives of their feature (i.e., input) space and target (i.e., output) space. Combining these two metrics we propose a Variation-Incentive Loss re-weighting method (VILoss) to optimize the gradient descent-based model training for regression analysis. We have conducted comprehensive experiments on both synthetic and real-world data sets. The results show significant improvement in the model quality (reduction in error by up to 11.9\%) when using VILoss as the loss criterion in training.
\end{abstract}

\section{Introduction}
\begin{figure}[htbp]
    \centering
	\includegraphics[width=200px]{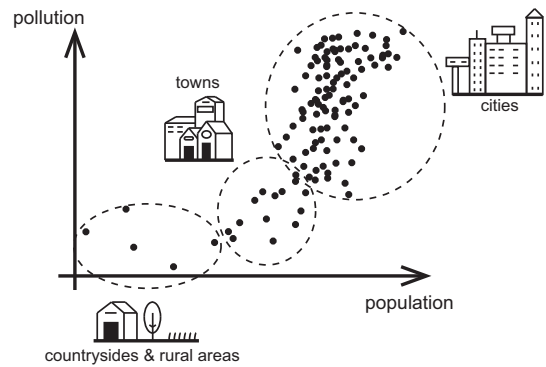}
	\caption{An example of population-pollution regression on a biased distribution of data collected from geographically distributed sensors that are not placed evenly across the country.}
	\label{fig:IoT_demo}
\end{figure}

Supervised machine learning is the centerpiece among all the achievements made in the era of artificial intelligence. This is more or less a result of the availability and diversity of real-world data sets annotated with labels. A well-known nature of supervised learning is its susceptibility to imbalanced or biased data distribution, which is very common for many reasons including the technical limitations in data collection, biased sampling methods and the inherent skewness in the data sources. Fig. \ref{fig:IoT_demo} shows an example of biased/skewed distribution of data collected from a massive number of sensors (e.g., $\mathrm{SO_2}$ sensors) that are used to indicate the air pollution in the corresponding residential areas. When we use the data to model the population-pollution relation using machine learning (e.g., gradient descent), the resulting model may learn the pattern well from the samples of densely populated areas (e.g., cities), but less effectively from those of rural areas. 

Although extensive efforts have been made to mitigate the impact of imbalanced data on classification tasks (e.g., \cite{imba_review1,imba_review2,imba_review3}), \textbf{none of the existing work can be directly applied to the regression problems}. For example, the common techniques such as re-sampling and re-weighting for handling the classification tasks cannot be applied to the pre-processing of the regression data or the training process of regression models. This is because the target space in a classification task is a set of discrete labels (classes) while that of a regression task consists of continuous numerical values.

%For the classification problems, the data imbalance is often referred to as the skewed distribution (e.g., long-tailed distribution \cite{CBLoss2019}) of labels in the target space. 
As a matter of fact, The data bias exists in both the feature (i.e., input) space and the target (i.e., output) space concerning a regression problem. 
%Fig. \ref{fig:skewness_demo} shows a simple case where the sampled data reside densely near the left boundary of the underlying function's feature range, but scarcely at the tail, resulting in the skewness in both its feature space (i.e., the \textit{x}-dimension values of the data are not evenly distributed on the X axis) and the target space (i.e., the Y axis). For the classification problems, data imbalance is referred to as the skewed distribution (e.g., long-tailed distribution \cite{CBLoss2019}) of labels in the target space. 
The data distribution can be skewed in the feature space due to the bias in data collection and the sourcing property of each feature. For example, the data samples may form a few clusters rather than scatter nicely over the entire feature space, which makes the distribution of the information largely uneven.

%In addition to the lack of studies in data skewness in the feature space, none of the existing re-balancing methods for classification are suitable for addressing the skewness problem in regression tasks. Specifically, the common techniques such as re-sampling \cite{re-sample1} and re-weighting \cite{re-weighting1} for handling the classification tasks cannot be applied to the pre-processing of regression data or the training process of regression models, since the target space in a regression task is not discrete by the labels naturally, but consists of continuous numerical values.

%% skewness demo
%\begin{figure}[htbp]
%    \centering
%    \includegraphics[width=210px]{skewness_demo.eps}
%    \caption{A demonstration of data skewness in a regression task with the dimensionality of feature space and target space both being 1.}
%    \label{fig:skewness_demo}
%\end{figure}

Biased data distribution may significantly degrade the efficacy of the regression analysis, which includes a wide range of learning tasks (e.g., anomaly detection and time series prediction) and practical applications (e.g., power estimation \cite{CAM2017, Lin2018}, price prediction, and event detection \cite{shi2018}). The success in dealing with imbalanced classes (in classification) inspires us to improve the quality of model training for regression problems such linear regression and numerical forecast. In this work, we adopt a novel approach to re-valuing the data samples according to their distributions in the feature (input) space and the target (output) space, and apply the approach to re-weight the loss during gradient descent based training.
%aim to improve the model training by re-weighting each data sample based on its learning value given a skewed regression data set. We first present a space partitioning method to determine the neighborhood of a sample, and use two simple variation-based metrics to characterize the uniqueness (i.e., how distinct the data sample is) and abnormality (i.e., how likely the data sample is an anomaly) of a data sample. We combine these two terms into a sample-wise weight and apply it to the base loss function in the gradient descent-based model training. By doing so, the data samples with more useful information will be able to make more contribution towards the final model, while outliers and redundant samples will be examined less. Thereby the influence of data skewness can be mitigated.
To the best of our knowledge, our study is the first to optimize the models for regression analysis by means of loss re-weighting. The key contributions of our work are outlined as follows:

\begin{itemize}
\item We propose to partition the feature space into a grid of \textit{cells}, and further introduce two metrics, uniqueness and abnormality, to indicate the data's learning values based on their variation.
\item We propose a loss re-weighting method (\textit{VILoss}) for optimizing the regression analysis and present an easy-to-implement method for determining its hyper-parameter through empirical studies.
\item We have conducted comprehensive experiments on both synthetic and public data sets; the results show a solid improvement in model accuracy when the proposed VILoss is used as the loss criterion.
\end{itemize}

%The rest of this paper is organized as follows. The related work on addressing data bias/imbalance is discussed in section \ref{secII}. In section \ref{secIII}, we introduce the notion of \textit{data cell} and define the uniqueness and abnormality of data samples. In Section \ref{secIV}, we integrate the proposed \textit{VILoss} into different base loss functions. In Section \ref{secV} the empirical studies are conducted to gain the insight into setting the appropriate values of the hyper-parameter in VILoss. We present experimental results in section \ref{secVI}. In section \ref{secVII} we conclude this paper.

\section{Related Work}
\label{secII}
The primary purpose of this work is to find an approach to optimizing model training for regression tasks. Nonetheless, it is worthwhile to discuss the popular methods for tackling the class imbalance problem. Through the discussions we can understand why the methods developed for classification problems cannot be applied to the regression problems we are targeting in this work.    

The most popular methodologies for addressing data imbalance/bias are re-sampling and re-weighting \cite{imba_review0}. Re-sampling is applied to the raw data set directly by adding repeated, interpolated or synthesized samples \cite{SMOTE2, over_sampling1} to the minority classes and/or removing a portion of samples in the majority classes \cite{under_sampling2}. A problem of the re-sampling method is that when the samples carrying the useful information are likely to be removed in undersampling, while the addition of redundant samples may introduce extra noise \cite{cons_resampling1} when conducting oversampling. 
Re-weighting is also commonly used to address data imbalance. Its fundamental idea is to assign different weights to different samples. Generally, a re-weighting method can be either model- and loss-agnostic \cite{re-weighting1, CBLoss2019} or error/loss-incentive \cite{error-incented1, focal17}. An example of the former is the class frequency-based re-weighting, whilst the later tends to assign larger weights to those "hard" samples that yields higher error.
%Similarly, Cao et al. \cite{loss_reweighting1} designed a label-distribution-aware margin (LDAM) loss and proposed to defer the re-weighting process at the end of the initial stage of training. The re-weighting methods can also be combined with the ensemble learning \cite{ensemble1,ensemble2,ensemble3} in order to handle the imbalanced class distribution by taking advantage of a group of base classifiers.

%Imbalanced data distribution is frequently met in the processing of sensor data, especially for the tasks such as novelty and anomaly detection \cite{shi2018}. In these cases, normal data samples significantly outnumber abnormal samples or outliers, making it very difficult for the model to learn and recognize abnormal patterns. Gao et al. \cite{Gao2016} figured that a main consequence of learning from imbalanced data is high false-alarm rate, and proposed an adaptive weighted imbalance learning method (AWELM). They further developed a fall detecting and monitoring application using multiple wearable eyeglasses and watches. The work boosted the performance of the classification model by using a two-stage detecting strategy and applying different weights to abnormal data in different stages. 

However, none of these existing methods can be applied to regression analysis, where the data are not affiliated to discrete class labels. Regression analysis entails a great number of practical applications such as numerical estimation \cite{Lin2018}, prediction \cite{CAM2017} and anomaly/novelty detection \cite{ADLTI2020}. Biased data distribution commonly exists in these tasks and could affect the quality of the models learned for regression analysis, which is hardly investigated in previous studies. 
%However, as common as imbalanced classification, the skewness may also exist in the regression data since the data distribution can be fairly uneven. As a result, training on such data sets may lead to an inferior model with the bias towards the pattern learned from the space where samples densely reside. 
%Liu and Chawla \cite{QMLearn11} proposed QMLearn, which aims to overcome the bias towards the majority classes in the classification problems by adopting the quadratic mean operation in the loss function. Theoretically, QMLearn can also be applied to regression tasks; but it is limited to handling left- or right-skewed distribution in the target space.

\section{Re-valuing Data with Variation}
\label{secIII}
The idea behind our re-weighting method is to differentiate each data sample by its learning value through gathering localized information on the data distribution (e.g., bias and outliers).

%For machine learning problems, a large volume of data is often considered to be very useful. However, how much a model can benefit from the data in practice is far more complicated than the data volume. For classification problems, Cui et al. \cite{CBLoss2019} find that the effective information contained in a class of samples is not proportional to its size because of "overlapping". Our studies show that the similar effect also applies to those data sets used for the regression problems -- the overlap of data in the feature space reduces the total information they carry. Consider a case where 80 percent of the samples are identical in terms of their feature values - the rest 20 percent may well contribute more to the established model when it is trained on the data. In addition, abnormal data samples can poison the resulting model of machine learning. This type of samples are sometimes called "outliers" as they may significantly deviate from the underlying patterns and the input-output relation represented by the data set.

We propose to re-value data samples by gauging each sample in terms of its uniqueness and (potential) abnormality. Uniqueness indicates how valuable a sample is for learning compared to the others while abnormality measures how likely it could be a deviant/outlier. The properties of uniqueness and abnormality are closely associated to the data variation in the feature space and the target space, respectively. 

We characterize uniqueness and abnormality of a sample in the context of its vicinity, which consists of a set of neighboring data samples. A straightforward method of finding a sample's neigbhorhood is to calculate and sort its distances to all others, and then select the closest \textit{k} samples. However, the time complexity is $O(n^2+n^2logn)$, which may result in unacceptably long pre-processing time for large data sets. In this work, we introduce \textit{Data Cell} as an approximate form of neighborhood. With the concept of the cell, the data can be examined efficiently in groups.

\subsection{Data Cell}
\label{sub:cell}
A data cell, in this paper, is defined as a logical neighborhood of data as a result of the dimension-wise partitioning in the feature space. For example, a three-dimensional feature space can be divided into $2^3$ cells with two divisions in each dimension. A cell can be formulated as: 

% cell definition in math
\begin{equation}
\label{eq:cell_ref}
    D_{p_1p_2...p_m} = \Big\{(\bsym{x}, \bsym{y}) \, | \, x^{(k)} \in p_k, \forall k=1,2,...,m\Big\}
\end{equation}
where $(\bsym{x}, \bsym{y})$ is a data sample (both $\bsym{x}$ and $\bsym{y}$ can be multi-dimensional), $x^{(k)}$ denotes the $k$-th feature and $m$ is the dimensionality of the feature space. The cell's subscript $p_1p_2...p_m$ is a string of numbers indexing the cell where $p_k$ stands for its projected index in dimension $k$. For the ease of the presentation, we refer to a cell $D_{p_1p_2...p_m}$ by the notation $D_p$ in the rest of this paper, where $p$ can be regarded as the rank of the cell after all cells are sorted in the ascending order by their index. The logic of binding samples to cells is similar to the space discretization when using the numerical method to solve equations, in which the continuous space is discretized into a grid of cells.

We use a hyper-parameter $\lambda \in \mathbb{Z}^+$ to control the number of divisions in each dimension and each division is of equal size. Given $n$ samples, the time complexity of partitioning the space into cells is $O(n)$ as we only need to traverse the data set once to assign each data sample to its corresponding cell. Given a raw feature space $\bsym{X}= \mathbb{R}^v$ with $v$ features, a full partitioning yields $\lambda^v$ cells. But in the case where $\bsym{X}$ is high-dimensional, we can select a subset $\bsym{X'} \subset \bsym{X}$ of features for space partitioning and loss weighting (still training the model with the full feature set $\bsym{X}$). After the feature space is divided, we can efficiently characterize the uniqueness and abnormality of all samples in the context of each cell. 
%$\lambda$ is the only hyper-parameter in our method. The bigger the $\lambda$ is, the smaller a cell (i.e., the neighborhood) becomes. However, a greater value of the hyperparameter $\lambda$ (i.e., each cell is smaller) does not necessarily lead to a better training outcome. This is because a smaller cell is more likely to contain fewer data samples and consequently contains less collective information to gauge samples' value (i.e., uniqueness and abnormality in our case) in the cell context. Hence an appropriate value of $\lambda$ plays an important role in our method for improving the quality of model training. We have conducted empirical studies on finding an appropriate value of $\lambda$, which is presented in section \ref{secV}. 

\subsection{Uniqueness: Feature Variation}
\begin{figure}[ht]
    \centering
    \includegraphics[width=230px]{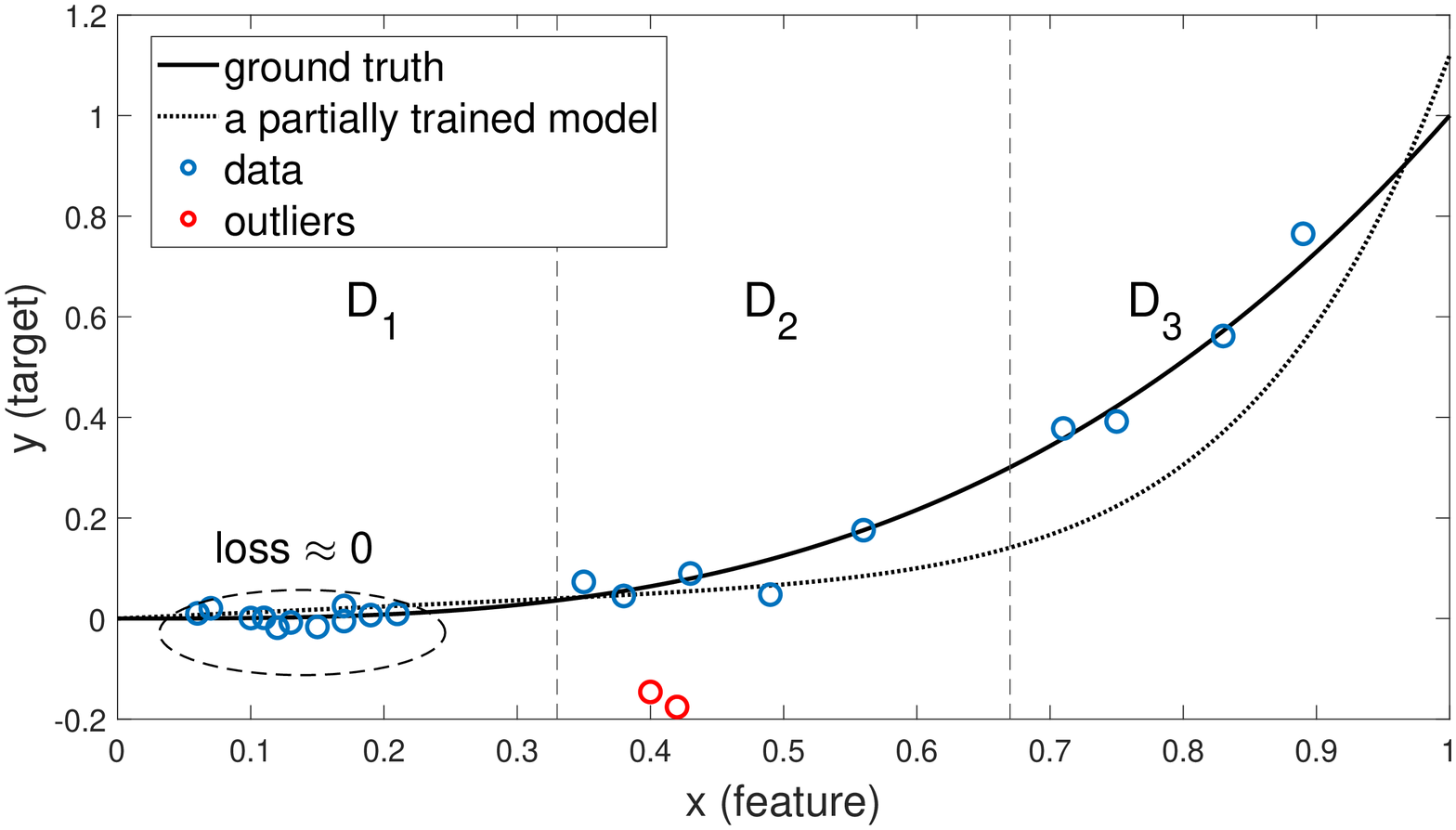}
    \caption{Illustration of data bias and its effect. Samples in cell $D_1$ are densely located and will not make much contribution to the further improvement of the model after the early stage of training. $D_2$ contains two abnormal samples that exhibit large deviations in the context of this cell.}
    \label{fig:demo_both}
\end{figure}

Intuitively, a unique sample is more likely to carry useful information for learning than a redundant one or one that overlaps with others. For example, assume a feature space shown in Fig. \ref{fig:demo_both}, which consists of two cells in which there exists a strong bias in cell $D_1$ where the samples reside within a small range and gather very closely together. When training a model on this data set, traditional training methods are likely to converge slowly as the model learns very little from the majority of samples in $D_1$.

% uniqueness demo

We measure how unique a sample is with respect to its features using a simple criterion based on variation. %Although it could be more precise to define the uniqueness for every single sample, we use the feature variation of a cell to represent the uniqueness of all the samples in it in order to reduce the time complexity as discussed above. 
Variation can be quantified statistically by the standard deviation and we generalize its formulation to fit the multi-dimensional feature space in the context of a cell using the Euclidean distance. Given a cell $D_p$, we define:

% sigma_x
\begin{equation}
\label{eq:sigma_x}
    \sigma_x(D_p) = \sqrt{\frac{1}{n_p} \sum_{(\boldsymbol{x,y}) \in D_p}
    	\| \bsym{x} - \bar{\bsym{x}}_{D_p} \|_2^2}
\end{equation}
where $\bar{\bsym{x}}_{D_p}$ is a vector of mean values over the corresponding elements of all $\bsym{x}$ in the cell $D_p$. %$\| \bsym{x} - \bar{\bsym{x}}_{D_p} \|_2$ denotes the Euclidean distance between $\bsym{x}$ and $\bar{\bsym{x}}_{D_p}$.

Using the generalized standard deviation, we can easily measure the variation of the data in the feature space and characterize their uniqueness. %Higher variation of the data in a cell means the higher diversity in terms of features, and thereby indicates a higher likelihood that the samples inside the cell are unique. As shown in Fig. \ref{fig:demo_both}, the variation of the feature (i.e., x-value) is much higher in $D_2$ than that in $D_1$, signifying that the samples are less overlapped in $D_2$ and potentially carry more useful information for learning. 
The samples with more distinct features (i.e., more unique) are deemed to carry more useful information for (further) optimizing the model compared to the less unique samples. Therefore, the uniqueness should be a relative value after considering all cells. We define the uniqueness $\mu$ of a cell $D_p$ as follows:  

% uniqueness cell
\begin{equation}
\label{eq:uniqueness_cell}
    \mu(D_p) = \frac{\sigma_x^2(D_p)}{\bar{\sigma}_x^2}
\end{equation}

% avg sigma_x
\begin{equation}
\label{eq:avg_sigma_x}
    \bar{\sigma}_x = \textrm{mean}\big(\sigma_x(D_p)\big), \; \forall D_p \in \bigcup_{|D_k| \neq 0}D_k
\end{equation}
where $\bar{\sigma}_x$ represents the average value of $\sigma_x(D_p)$ over all non-empty cells (The empty cells are excluded in the calculation). All samples in a cell share the uniqueness value of this cell: 

%Since we use the uniqueness of a cell to represent the uniqueness of every sample in that cell, the uniqueness value of a sample $(\bsym{x,y})$ is formulated as follows:

% uniqueness sample
\begin{equation}
\label{eq:uniqueness_sample}
    \mu(\bsym{x,y}) = \mu(D_p), \; \forall (\bsym{x,y}) \in D_p
\end{equation}

\subsection{Abnormality: Target Variation}
It is very common to have the corrupted data in a large data set, which deviate significantly from the actual distribution of the underlying function, and are poisonous to the training process. We judge the abnormality of the samples from the perspective of the target space. Given a cell of data, a large variation of intra-cell samples in the target space usually means the inconsistency, since in general the neighboring samples have similar feature values and thus are expected to have relatively similar target values. %The existence of abnormal data increases the uncertainty in the target space (i.e., output space) and consequently misleads the learning process. 
For example, the two outliers (highlighted red) among the data shown in Fig. \ref{fig:demo_both} could mislead the model because they yield high loss and gradients as their Y values deviate significantly from the true distribution. 
%% uniqueness demo
%\begin{figure}[htbp]
%    \centering
%    \includegraphics[width=210px]{abnormality_demo_div4.eps}
%    \caption{A demonstration of the negative impact of the abnormal data on the learning result of a regression model (the dashed line); Abnormal samples are marked as red circles.}
%    \label{fig:abnormality_demo}
%\end{figure}
Therefore, we propose to gauge the abnormality of a given sample by it target value in the context of its cell. Taking the average target value $\bar{\bsym{y}}$ in the cell as a reference, the deviation of a sample $(\bsym{x,y})$ in the target space can be measured using the $p$-norm distances, i.e., $\| \bsym{y} - \bar{\bsym{y}} \|_p$. In this paper we use L1 norm ($p=1$) and the Euclidean norm ($p=2$). We define the target variation of cell $D_p$ as:

% sigma_y
\begin{equation}
\label{eq:sigma_y}
    \sigma_y(D_p) = \sqrt{\frac{1}{n_p} \sum_{(\bsym{x,y}) \in D_p} \|\bsym{y} - \bar{\bsym{y}}_{D_p} \|_2^2}
\end{equation}
where $n_p$ denotes the size (i.e., the number of samples inside) of $D_p$ and $\bar{\bsym{y}}_{D_p}$ denotes the average $\bsym{y}$ value of the samples in the cell. Using $\sigma_y(D_p)$ as the reference, given a sample $(\bsym{x,y})$ in cell $D_p$, we quantify its level of abnormality $\gamma$ using two forms (L1 and L2 forms):

% L1 risk
\begin{equation}
\label{eq:L1_risk}
    \gamma_{L1}(\bsym{x,y}) = \frac{| \bsym{y}- \bar{\bsym{y}}_{D_p} |}{\sigma_y(D_p)} \, , \, (\bsym{x,y}) \in D_p
\end{equation}

% L2 risk
\begin{equation}
\label{eq:L2_risk}
    \gamma_{L2}(\bsym{x,y}) = \frac{\| \bsym{y} - \bar{\bsym{y}}_{D_p} \|_2^2}{\sigma^2_y(D_p)} \, , \, (\bsym{x,y}) \in D_p
\end{equation}
%where $\gamma_{L1}(\bsym{x,y})$ and $\gamma_{L2}(\bsym{x,y})$ denote the level of abnormality of the sample $(\bsym{x,y})$ in L1 and L2 forms, respectively. The definition of the abnormality reflects how significantly a sample deviates from its feature-space neighbors in the target space. The higher the abnormality level of the sample is, the more likely it is an outlier that misleads the model training.

\section{Variation-Incentive Loss}
\label{secIV}
%Although loss re-weighting methods (e.g., \cite{CBLoss2019}\cite{QMLearn11}) have shown their effectiveness in promoting the performance of the classification models trained on imbalanced data sets with gradient descent, it is challenging to apply them to the regression problems. In this paper, we propose a novel re-weighting method to improve the model learning on skewed regression data sets. We propose a method to weight commonly used loss functions based on the learning values of the samples. 
We re-value the data samples to optimize the learning by taking into account their uniqueness and abnormality properties introduced in section \ref{secIII}. Specifically, we assign higher weights to the samples that exhibit strong uniqueness and low abnormality. Given a sample $(\bsym{x,y})$, a predicted output $\hat{\bsym{y}}$ and a base loss function $\mathcal{L}$ (for the regression task), the proposed Variation-Incentive Loss (\textit{VILoss}) is formulated in (\ref{eq:VILoss}):

% VILoss
\begin{equation}
\label{eq:VILoss}
    \textrm{VILoss}(\bsym{x,y},\hat{\bsym{y}}) = \frac{\mu(\bsym{x,y})}{1+\gamma(\bsym{x,y})} \mathcal{L}(\hat{\bsym{y}},\bsym{y}) 
\end{equation}
%where $\mu(\bsym{x,y})$ denotes the uniqueness of the data sample defined in (\ref{eq:uniqueness_sample}), $\gamma(\bsym{x,y})$ is the abnormality of the sample, which can take two different forms: $\gamma_{L1}(\bsym{x,y})$ and $\gamma_{L2}(\bsym{x,y})$ defined in (\ref{eq:L1_risk}) and (\ref{eq:L2_risk}), respectively. The base loss function $\mathcal{L}$ takes two arguments (i.e., $\bsym{y}$ and $\hat{\bsym{y}}$) and computes their difference as the error. One difference of VILoss from the conventional loss function is that it includes the features (i.e., $\bsym{x}$) of the sample, which is needed to determine which cell the sample belongs to.

In (\ref{eq:VILoss}), we weight the loss of a sample positively proportional to its uniqueness $\mu(\bsym{x,y})$ and inversely proportional to its abnormality $\gamma(\bsym{x,y})$ plus one. By doing so we can modulate how much attention the sample obtains in training. We use $1+\gamma(\bsym{x,y})$ in the denominator to avoid the extreme values in case $\gamma(\bsym{x,y}) \approx 0$. Note that since the weight applied to the loss is independent of the predictions made by the model and the model parameters, we can easily calculate the gradient of a sample $(\bsym{x,y})$ with the VILoss as the loss function:

% VILoss gradient
\begin{align}
\label{eq:VILoss_grad}
  \nabla_{\bsym{w}}(\textrm{VILoss}(\bsym{x,y},\hat{\bsym{y}})) 
    &= \frac{\partial \textrm{VILoss}(\bsym{x,y},\hat{\bsym{y}})}{\partial \bsym{w}} \nonumber \\
    &= \frac{\mu(\bsym{x,y})}{1+\gamma(\bsym{x,y})} \frac{\partial \mathcal{L}(\hat{\bsym{y}},\bsym{y})}{\partial\bsym{w}} 
\end{align}
where $\bsym{w}$ denotes the model parameters for training. %With VILoss, the gradient of each sample is modulated based on its learning value (according to its uniqueness and abnormality), which enables the samples with higher learning value to contribute more to the update of the model parameters and consequently leads to a boost in the model's quality.

%tasks which produce models that output numerical values. The main reason why VILoss is not suitable for most of the classification problems is that the class labels are discrete. When computing the sample's abnormality, we need to calculate the distance between the value of a sample and the average value of the samples in the target space. The distance between the discrete classes does not make sense. 

%Nevertheless, the design of VILoss makes it applicable to training the logistic regression models for binary classification. This is because in this case we can establish the cell average $\bar{y}_{D_p}$ arithmetically in the target space where the label $y$ takes the value of 0 or 1 while $0 \leq \bar{y}_{D_p} \leq 1$. Therefore, the proposed VILoss method can be applied to both regression and binary classification problems. Next, we present how VILoss is applied to the most commonly used loss functions for regression (subsection \ref{sub:VILoss_reg}) and logistic regression based binary classification problems (subsection \ref{sub:VILoss_log}). 

\bigskip
\noindent \textit{VILoss for Numerical Regression}

VILoss is designed for optimizing models for regression analysis that can be formulated as: $f: \mathbb{R}^m \rightarrow \mathbb{R}^v$, where $m$ and $v$ are the dimensionality of the feature space and the target space, respectively. With the prevalence of the gradient descent-based training methods, the commonly used loss functions for regression include Mean Square Error (MSE) loss (a.k.a., L2 loss) and Huber loss (a.k.a., smooth L1 loss).%these three loss functions are most commonly used for regression tasks: Mean Absolute Error (MAE) loss (a.k.a., L1 loss), Mean Square Error (MSE) loss (a.k.a., L2 loss) and Huber loss (a.k.a., smooth L1 loss). Compared to other two, L1 loss is barely used in training modern machine learning models because it is not differentiable at the origin and yields large gradients even for small losses, which makes it not suitable for complicated models such as neural nets. Hence, we focus on applying the proposed re-weighting method to L2 loss and Huber loss.

Given a sample $(\bsym{x,y})$ and the corresponding output $\hat{\bsym{y}}$ predicted by a regression model, we have $VILoss_{MSE}$ using MSE as the base loss:

% VILoss_MSE
\begin{equation}
\label{eq:VILoss_MSE}
    \textrm{VILoss}_\textrm{MSE}(\bsym{x,y}, \hat{\bsym{y}}) = 
      \frac{\mu(\bsym{x,y})}{1+\gamma(\bsym{x,y})} \frac{1}{v} \sum_{i=1}^v (\hat{y}^{(i)}-y^{(i)})^2 
\end{equation}
where $v$ is the number of dimensions in the target space and $y^{(i)}$ is the element in the $i$-th dimension. In a similar way we can apply our re-weighting method to higher-ordered loss functions like the Least Quartic Regression (LQR) loss \cite{LQR2020}, which uses a quartic function of error (i.e., $(\hat{y}^{(i)}-y^{(i)})^4$) that yields significantly higher losses on difficult samples.

Huber loss is a piece-wise function that is linear when the absolute error $|\hat{y}^{(i)}-y^{(i)}|$ is above a threshold $\delta$ and is smoothed to a quadratic form otherwise. Applying our re-weighting method, we define $\textrm{VILoss}_\textrm{Huber}$ as:

% VILoss_Huber
\begin{equation}
\label{eq:VILoss_Huber}
    \textrm{VILoss}_\textrm{Huber}(\bsym{x,y}, \hat{\bsym{y}}) = 
      \frac{\mu(\bsym{x,y})}{1+\gamma(\bsym{x,y})} \frac{1}{v} \sum_{i=1}^v z_i(\hat{\bsym{y}},\bsym{y})
\end{equation}

% the z of VILoss_Huber
\begin{equation}
\label{eq:VILoss_Huber_z}
    z_i(\hat{\bsym{y}},\bsym{y}) = 
    \begin{cases}
     \frac{1}{2} (\hat{y}^{(i)}-y^{(i)})^2 & \textrm{if }|\hat{y}^{(i)}-y^{(i)}|<\delta,\\
     \delta |\hat{y}^{(i)}-y^{(i)}| - \frac{1}{2} \delta^2 & \textrm{otherwise.}   
    \end{cases}
\end{equation}
where $z_i(\hat{\bsym{y}},\bsym{y})$ is the smooth L1 error for the $i$-th output dimension, and $\delta$ is a hyper-parameter for Huber loss and is often set to 1.0 empirically.

\bigskip
\noindent \textit{VILoss for Logistic Regression}

VILoss is designed for the regression tasks which requires the models to output numerical values. Nevertheless, the design of VILoss makes it applicable to training logistic regression models for binary classification. This is because in this case we can establish the cell average $\bar{y}_{D_p}$ arithmetically in the target space where the label $y$ takes the value of 0 or 1. %Therefore, the proposed VILoss method can be applied to both regression and binary classification problems. Next, we present how VILoss is applied to the most commonly used loss functions for regression and logistic regression based binary classification problems.
Logistic regression models here refer to any form of models (e.g., linear regression, recurrent neural networks) with a sigmoid function applied to its output. As a result, the final output $\hat{y}$ for binary classification is a scalar of probability. In this paper, we use Binary Cross Entropy (BCE) as the base function of loss:

% VILoss BCE
\begin{equation}
\label{eq:VILoss_BCE}
    \textrm{VILoss}_\textrm{BCE}(\bsym{x},y, \hat{y}) = 
      - \frac{\mu(\bsym{x},y)}{1+\gamma(\bsym{x},y)} [y\textrm{log}\hat{y} + (1-y)\textrm{log}(1-\hat{y})]
\end{equation}
where $y$ is the class label for the sample $(\bsym{x},y)$ taking either 0 or 1, and $\hat{y}$ denotes the prediction (in the range $(0,1)$) made by the model with a logistic output layer.

\section{Study of the Hyper-parameter $\lambda$}
\label{secV}
The hyper-parameter $\lambda$ (which determines the number of cells) plays an important role in the training outcome as we need to partition the feature space appropriately to differentiate the value of samples. On the one hand, as $\lambda$ increases, the partitioning granularity gets finer (i.e., more cells), which is good because we use the cell uniqueness to represent the uniqueness of all samples in this cell. On the other hand, as $\lambda$ increases, the cells become sparser, which means more information is lost regarding the local distribution of samples.
To find the optimal value of $\lambda$ before training, we define a simple metric called \textit{Localized Deviation} (\textit{LD}) to indicate if the feature space is appropriately partitioned given a set of data. \textit{LD} is defined as the sum of $\sigma_x(D_p)$ concerning all the non-empty cells (i.e., $D_p \neq \emptyset$) as a result of partitioning:

% LD
\begin{equation}
\label{eq:LD}
    LD = \sum_{|D_p| \neq 0} \sigma_x(D_p)
\end{equation}

\begin{figure}[htbp]
    \centering
    \includegraphics[width=235px]{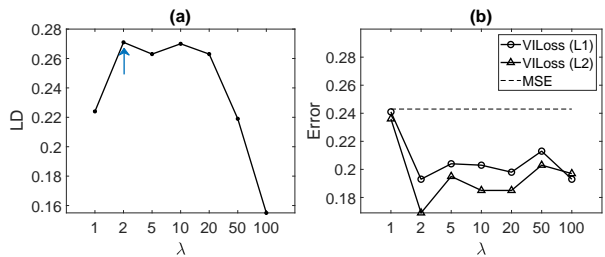}
    \caption{Test results of the models trained on the Synth-1D data set to show the correlation between (a) the value of LD, and (b) the best error using different $\lambda$.}
    \label{fig:LD_1D}
\end{figure}

\begin{figure}[htbp]
    \centering
    \includegraphics[width=235px]{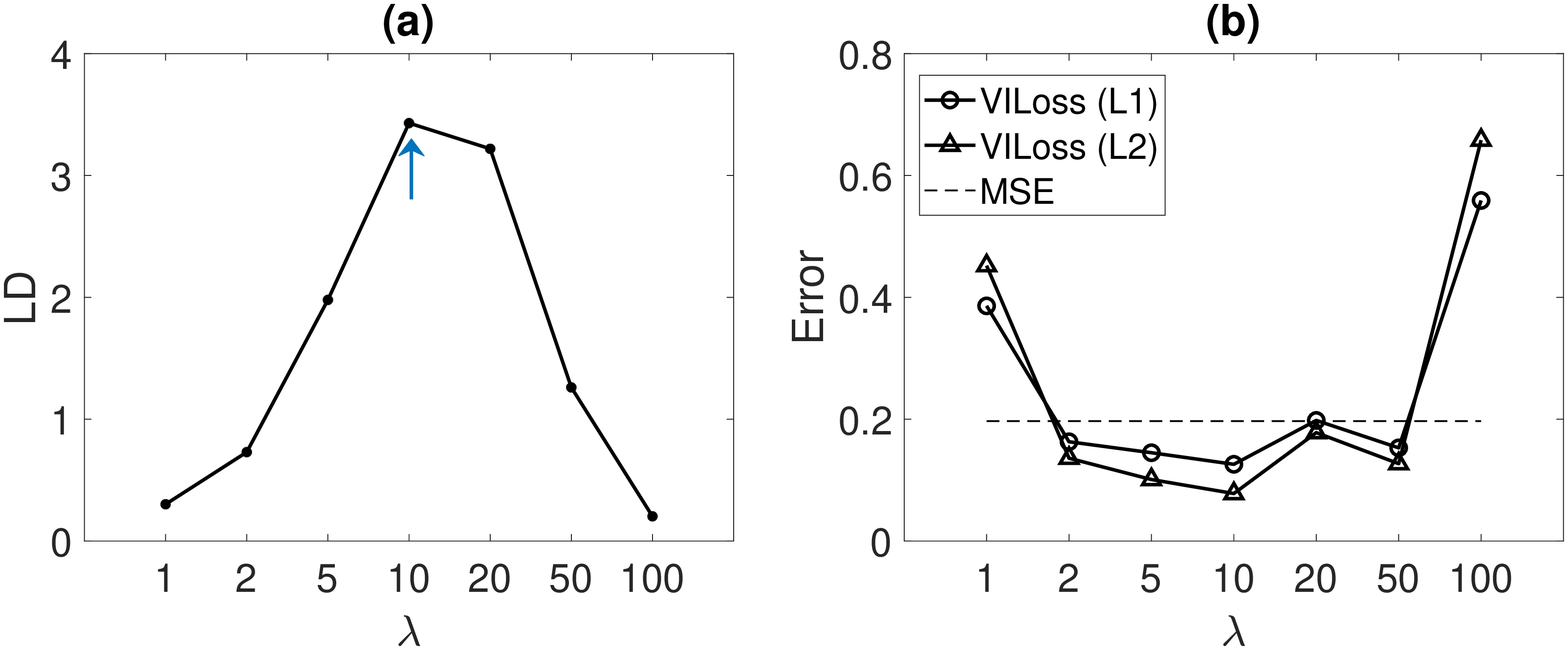}
    \caption{Test results of the models trained on the Synth-2D data set to show the correlation between (a) the value of LD, and (b) the best error using different $\lambda$.}
    \label{fig:LD_2D}
\end{figure}

We use the following empirical studies to show that the designed \textit{LD} can be used to guide the selection of the optimal $\lambda$. We conducted the empirical studies on two synthetic data sets: Synth-1D and Synth-2D. Each set is generated from a polynomial data model with  non-biased Gaussian noise (simulating normal variation) and a certain portion of abnormal data (simulating data corruption). We trained the models of varied complexity on these two data sets to observe the correlation between $\lambda$, \textit{LD} and the model's quality (measured in Mean Absolute Percentage Error, MAPE). Under each setting we report the lowest error achieved from all candidate models. Fig. \ref{fig:LD_1D} and Fig. \ref{fig:LD_2D} plot the results extracted from the full statistics in Tables \ref{tab:synth_1D_full} and \ref{tab:synth_2D_full} in Appendix \ref{apx}.

% LD synth-2D

On Synth-1D, VILoss outperformed MSE under all settings, and the lowest error is achieved by setting $\lambda$ to 2, which also yields the maximum value of \textit{LD} (Fig. \ref{fig:LD_1D}). The case is different on Synth-2D wherein the value of \textit{LD} is maximized at $\lambda=10$ (Fig. \ref{fig:LD_2D}) while VILoss also achieves the biggest reduction of error, which means $10^2$ cells represent the best partitioning of this data set. In both cases we find \textit{LD} is a strong indicator of the effectiveness of our method. Based on these observations, we are able to determine an appropriate value of $\lambda$ without actually running the training process, but by running this procedure instead: 1) Select a group of candidate values of $\lambda$; 2) Calculate the value of \textit{LD} for each candidate $\lambda$ by using (\ref{eq:LD}); 3) Choose the value of $\lambda$ that maximizes \textit{LD}.

\section{Evaluation}
\label{secVI}
\subsection{Experiment Setup}
\begin{figure}[htbp]
    \centering
	\includegraphics[width=220px]{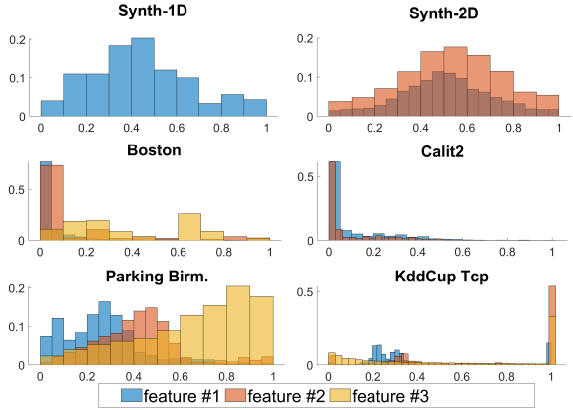}
	\caption{Biased feature distribution of the data used in our experiment. For the data sets with dimensions $>$ 3, only three features are shown.}
	\label{fig:dataset_features}
\end{figure}

We conducted extensive experiments by running a diversity of regression tasks on both synthetic and real-world data sets. As summarized in Table \ref{tab:exp_setup}, we chose four public data sets that correspond to four typical application scenarios including house price regression (\textit{numerical value estimation}), parking vacancy prediction (\textit{time series prediction}), cyber intrusion detection (\textit{anomaly detection}) and event detection (\textit{time series anomaly detection}). For each set, 70\% of the data are used for training and 30\% for test. All selected data sets (including our Synth-1D and Synth-2D) are more or less biased in data distribution. The data are normalized on both features and targets. In Fig. \ref{fig:dataset_features} we show the feature distributions of these data sets among which only our synthetic data follow clear patterns of Gaussian distribution.

% experiment setup table
\begin{table}[htbp]
\centering
\caption{Experimental setup: data sets and models}
\begin{tabular}{ l c l } 
 \hline
 Data set 	  	 		&Dimensions 	&Model 				\\ 
 \hline
 Synth-1D				&1+1			&polynomial			\\
 Synth-2D	 			&2+1			&polynomial			\\
 Boston  				&13+1			&linear regression	\\
 Parking Birm.	 		&5+2			&LSTM				\\
 KDDCup'99		 		&41+1			&logistic regresion	\\
 CalIt2 				&2+1			&logistic RNN		\\
 \hline
\end{tabular}
\label{tab:exp_setup}
\end{table}

%% [Algorithm 1: Searching for optimal number of divisions per dimension]
%\begin{algorithm}[ht] 
%\caption{xxx}
%  \DontPrintSemicolon
%  \SetKwInOut{Input}{Input}
%  \SetKwInOut{Output}{Output}
%  \SetAlgoLined
%  \Input{}
%  \Output{clients to pick $P(t)$}
%  % use \rm before each statement containing text
%  return $P(t)$
%\label{algo:yyy}
%\end{algorithm}

\subsection{Experimental Results}
We applied our re-weighting method to model training using MSE, Huber and the Least Quartic Regression (LQR) loss \cite{LQR2020} as the base loss. MSE loss and Huber loss are the standard criteria for regression analysis while the LQR loss is designed for learning from the fat-tailed financial data that exhibit strong skewness or kurtosis. We also evaluated our method with the logistic regression tasks by using the popular BCE as the base loss. Our method is compared against the base losses in terms of the model's performance in tests. %Maximum number of epochs is set to 100 (sufficient for all the models in our experiments to converge) with early-stopping enabled to prevent over-fitting. 
All the results reported in the figures are collected under the optimal training settings (e.g., batch size, learning rate and time step).

\noindent \emph{Numerical Regression}

We used Boston (housing) and the Parking Birmingham data sets from the UCI repository for numerical regression tasks. The Boston housing\footnote{\url{https://www.cs.toronto.edu/~delve/data/boston/bostonDetail.html}} data set contains the information collected by the U.S. Census Service with 14 attributes in total of which the median house price is to be predicted. We use a linear regression model for this task. Parking Birmingham\footnote{\url{https://archive.ics.uci.edu/ml/datasets/Parking+Birmingham}} is a time-series data set collected by the city council. We extracted two months of the data (from Oct. 4 to Dec. 9, 2016) from the records of five car parks in the city of Birmingham and use them to predict the vacancies in two other car parks (Broad Street and Bull Ring) in the same city. On this data set we trained and tested two forms of LSTM neural nets with one and two stacked hidden layers, respectively. The size of hidden layer is set to 10.
Apart from the real-world data sets, the experimental results on synthetic data sets are also included for reference. The errors are reported in two metrics: mean absolute percentage error (MAPE) and mean absolute error (MAE). 

% table:synth_part
\begin{table}[htbp]
\centering
\caption{Experimental results on synth-1D and synth-2D data sets using polynomial regression models. The optimal $\lambda^*$ for our loss function on the two data sets are 2 and 10, respectively.}
\begin{tabular}{ l l l l l } 
 \hline
 			 	  	& \multicolumn{2}{c}{Synth-1D}	& \multicolumn{2}{c}{Synth-2D}	\\ 
 Loss function				& MAPE	& MAE	& MAPE	& MAE 	\\ 
 \hline
 MSE loss					& 0.243	& 0.018				& 0.197	& 0.015	\\ 
 $\mathrm{VILoss_{MSE}}$(L1)& 0.193	& 0.015			& 0.126	& 0.015	\\
 $\mathrm{VILoss_{MSE}}$(L2)& 0.169	& \textbf{0.012}& \textbf{0.078}& \textbf{0.010}\\
 Huber loss					& 0.232	& 0.018			& 0.228	& 0.018	\\
 $\mathrm{VILoss_{Huber}}$(L1)& 0.158	& 0.017		& 0.163	& 0.020	\\
 $\mathrm{VILoss_{Huber}}$(L2)& \textbf{0.155}& 0.019	& 0.123	& 0.019	\\
 LQR loss					& 0.261	& 0.058			& 0.336	& 0.040	\\
 $\mathrm{VILoss_{LQR}}$(L1)& 0.161	& 0.013			& 0.312	& 0.033	\\
 $\mathrm{VILoss_{LQR}}$(L2)& 0.180	& 0.016			& 0.361	& 0.047	\\
 \hline
\end{tabular}
\label{tab:synth_part}
\end{table}

% table:Boston_Birm
\begin{table*}[htbp]
\centering
\caption{Experimental results on Boston Housing and Parking Birmingham data sets using a linear regression model and an LSTM neural net, respectively. The optimal $\lambda^*$ for our loss function on the two data sets are 4 and 10, respectively.}
\begin{tabular}{ l l l l l l l} 
 \hline
 	& \multicolumn{2}{c}{Boston Housing}	& \multicolumn{4}{c}{Parking Birmingham}				 \\ 
 Model& \multicolumn{2}{c}{Linear regression}	& \multicolumn{2}{c}{LSTM-1} & \multicolumn{2}{c}{LSTM-2}\\ 
 Loss function		& MAPE	& MAE	&  MAPE	& MAE 	&  MAPE	& MAE\\ 
 \hline
 MSE loss						 & 0.326& 4.650			& 0.272& 264.4			& 0.301& 295.6\\ 
 $\mathrm{VILoss_{MSE}}$(L1)	 & 0.279& 4.752			& 0.188& 177.9			& 0.207& 232.2\\ 
 $\mathrm{VILoss_{MSE}}$(L2)	 & 0.277& 4.668			& 0.189& 180.7			& 0.200	&227.3\\ 
 Huber loss						 & 0.315& 4.562			& 0.258& 249.6			& 0.280& 267.0\\ 
 $\mathrm{VILoss_{Huber}}$(L1)& \textbf{0.274}& \textbf{4.453}	& \textbf{0.177}& \textbf{173.9}& 0.231& 263.1\\ 
 $\mathrm{VILoss_{Huber}}$(L2)& 0.276& 4.733			& 0.180	& 177.5			& 0.235	& 266.2\\ 
 LQR loss						 & 0.350& 5.028			& 0.233	& 258.2			& 0.215 & 269.3	\\ 
 $\mathrm{VILoss_{LQR}}$(L1)	 & 0.330& 5.038			& 0.207	& 247.4			& 0.194	& \textbf{214.8}\\ 
 $\mathrm{VILoss_{LQR}}$(L2)	 & 0.330& 5.113			& 0.213	& 250.8			& \textbf{0.192}& 220.5\\ 
 \hline
\end{tabular}
\label{tab:Boston_Birm}
\end{table*}

\begin{figure}[htbp]
    \centering
	\includegraphics[width=235px]{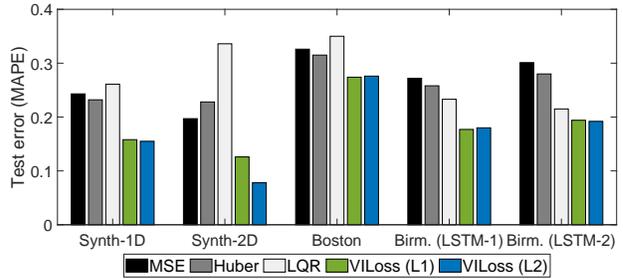}
	\caption{Comparing the test error of the models trained with MSE Loss, Huber Loss, LQR Loss and VILoss on different data sets. Note that two different model structures (i.e., one-layer LSTM and two-layer LSTM) are used for the Parking Birmingham data set.}
	\label{fig:bars_numreg}
\end{figure}

In Table \ref{tab:synth_part}, MAE and MAPE are reported by selecting the best polynomial models in each test (see Appendix A for details). On both synthetic data sets, the models are enhanced in performance (i.e., error reduction) when trained with our VIloss. $\mathrm{VILoss_{MSE}}$ in the L2 form achieved the best performance, reducing the relative error by 7.4\% and 11.9\% on Synth-1D and Synth-2D, respectively.

The test results on the Boston Housing and Parking Birmingham datasets are summarized in Table \ref{tab:Boston_Birm}. On Boston Housing, the model quality is improved by 4-5\% when VILoss is applied. When testing on the Parking Birmingham dataset, we also observed a significant performance gain especially for the stacked LSTM model (i.e., LSTM-2) with about 10\% error reduction compared to that trained with the MSE loss. Combining Tables \ref{tab:synth_part} and \ref{tab:Boston_Birm}, Fig. \ref{fig:bars_numreg} provides the comparison in terms of test error between VILoss and the baseline losses (MSE, Huber and LQR). From the figure we can observe a solid improvement in the accuracy (i.e., lower test error) of different models when trained using VILoss on both synthetic data sets and real-world data sets. %Experimental results on the Parking Birmingham data set also show that our loss re-weighting method is effective for optimizing the training of single- and multi-layer recurrent neural networks.

\noindent \emph{Logistic Regression}

Our approach can also be applied to train logistic regression models for binary classification. %We trained and tested a linear logistic regression model on the KDDCup'99 SA data set, and a vanilla RNN model with logistic output layer on CalIt2. 
We extracted a subset of the KDDCup'99 data set\footnote{\url{https://scikit-learn.org/stable/datasets/index.html\# kddcup99-dataset}} to include the TCP records only, and %The resulting subset size is 190065 entries. 
built a logistic regression model for predicting network intrusions (labelled 1 if true, 0 otherwise). For this data set, we weight the samples based on three of the features ('src\_bytes', 'dst\_bytes' and 'dst\_host\_count'). The CalIt2 data set\footnote{\url{https://archive.ics.uci.edu/ml/datasets/CalIt2+Building+People+Counts}} from UCI and is a time-series data set containing the time stamps and two data streams (in-flow and out-flow) of a department building where the events take place occasionally and need to be detected. We merged raw records hourly and built an RNN model with sigmoid output. Table \ref{tab:kddcup_calit2} shows the comprehensive test results of these two models with several classification metrics.

% table:synth_part
\begin{table}[htbp]
\centering
\caption{Experimental results on KDDCup'99 and CalIt2 data sets in terms of accuracy, precision, recall and F1-score. The optimal $\lambda^*$ for VILoss on the two data sets are 10 and 5, respectively.}
\begin{tabular}{ l l l l l} 
 \hline
 			 	  			& \multicolumn{4}{c}{KDDCup'99 TCP}						\\ 
 Loss function				& acc	& prec	& recall	& f1-score 	\\ 
 \hline
 BCE loss							& 0.927		& 0.921		& \textbf{0.995}		& 0.957		\\ 
 $\mathrm{VILoss_{BCE}}$(L1)		& \textbf{0.939}& \textbf{0.934}& 0.994		& \textbf{0.964}	\\ 
 $\mathrm{VILoss_{BCE}}$(L2)		& 0.938		& 0.933		& 0.994		& 0.963		\\ 
 \hline
 			 	  			& \multicolumn{4}{c}{CalIt2}									\\
 BCE loss							& 0.960		& 0.619		& 0.406		& 0.491		\\ 
 $\mathrm{VILoss_{BCE}}$(L1)		& 0.965		& 0.722		& \textbf{0.406}& \textbf{0.520}	\\ 
 $\mathrm{VILoss_{BCE}}$(L2)		& \textbf{0.966}& \textbf{0.800}		& 0.375		& 0.511		\\
 \hline
\end{tabular}
\label{tab:kddcup_calit2}
\end{table}

\begin{figure}[htbp]
    \centering
	\includegraphics[width=220px]{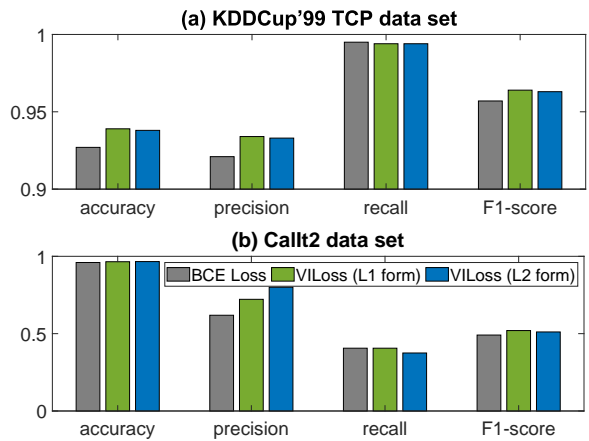}
	\caption{Comparing the accuracy, precision, recall and F1-score of the resulting models when trained with Binary Cross Entropy Loss and VILoss on (a) KDDCup'99 TCP data set (for anomaly detection) and (b) CalIt2 data set (for time series event detection).}
	\label{fig:bars_logireg}
\end{figure}

From Table \ref{tab:kddcup_calit2} we can observe the high recall rates for all the training criteria on KDDCup'99 TCP, but using VILoss improves the model's accuracy, precision and F1-score. This is because VILoss pays more attention to those uncharacteristic negative samples (with some degree of deviation from the negative majority) and thus reduces false positive decisions (false alarm rate). Event detection is much more difficult on the time series data CalIt2 even with recurrent layers. Some events (e.g., small conferences) do not increase the in-/out-flow of the building significantly, resulting in a relatively low recall rate. In this case, our improvement in recall rate is marginal. However, from Table \ref{tab:kddcup_calit2} we can observe a notable improvement in precision by roughly 10-20\% over the baseline and the comprehensive performance gain is about 3\%, as reflected by the f1-score. Fig. \ref{fig:bars_logireg} summarizes the experimental results on the KDDCup'99 TCP and CalIt2 data sets. It is shown that VILoss can effectively optimize the training of both traditional logistic regression models and recurrent neural nets with logistic output.

\section{Conclusion}
\label{secVII}
Biased data distribution commonly exists in all kinds of real-world data sets and is likely to deteriorate the models learned from these data. However, existing studies hardly pay attention to this problem for the regression analysis. In this paper, we present an approach to quantifying the uniqueness and abnormality of the samples in a skewed and biased data distribution. Combining the two metrics we propose a loss re-weighting method that can be applied to different base loss functions (e.g., MSE, Huber, LQR and BCE Loss) for the regression purpose. The experimental results on various models and data sets show a significant gain in model performance when using VILoss in training.

%In future we plan to extend our work to make further investigation into the impact of data skewness in a quantitative manner and study how to optimize our re-weighting method by transforming the raw feature space and reducing its dimensionality.

%\section*{Acknowledgement}
%This work is partially supported by Worldwide Byte Security Co. LTD, the Guangzhou Science and Technology Program key projects (Grant No. 202007040002), Guangzhou Development Zone Science and Technology project (Grant No. 2018GH17), Guangdong project (Grant No. 2018B030325002), International Cooperation Project of the Department of Science and Technology of Henan Province (Grant No. 172102410065) and the Frontier Interdisciplinary Project of Zhengzhou University (Grant No. XKZDQY202010).

% bib
\bibliographystyle{plain}
\bibliography{refs}

\newpage
\appendix
\section{Experiment details for parameter study}
\label{apx}

The ground truth of the data is $y=x^6+0.3$ for Synth-1D, and $y=-x_1+x_2^6+x_3^3+0.3$ for Synth-2D (here $x_1$ and $x_2$ represent the two features). The number of samples generated for these two synthetic data sets is 300 and 1000, respectively. 70 percent of each data set are used for training and 30 percent for test. On each data set we trained three polynomial models: a third-order polynomial model (which is under-parameterized because the degree of the ground truth model is 6), a sixth-order polynomial model (i.e., properly parameterized) and a tenth-order polynomial model (i.e., over-parameterized). The MSE Loss is used as the baseline. The batch size is set to 1 and 5 for Synth-1D and Synth-2D, respectively. We use MAPE (Mean Absolute Percentage Error) as the error metric. Results are summarized in Tables \ref{tab:synth_1D_full} and \ref{tab:synth_2D_full}.

We gather the results and investigate the correlation between the effectiveness of the proposed VILoss and the \textit{LD} metric we define in (\ref{eq:LD}). The best result is selected for each loss function (i.e., MSE, VILoss L1-form and VILoss L2-form) with each specified value of $\lambda$ from 2 to 100. We use MAPE as the error metric.

% table:synth_1D_full
\begin{table}[ht]
\centering
\caption{Full test results of the error of models trained with VILoss and the base loss on Synth-1D. Poly-3 and Poly-10 models are used to show the cases when under-parameterized and over-parameterized models are trained on the data set, respectively.}
\begin{tabular}{ l l l l } 
 \hline
 	 & \multicolumn{3}{c}{test error of models}										\\
 	 						& Poly-3			& Poly-6 			& Poly-10		\\ 
 \hline
 MSE loss					& 0.447				& 0.315				& 0.243			\\
 \hline
 \multicolumn{4}{c}{$\lambda=1$ (no partitioning)}									\\
 $\mathrm{VILoss_{MSE}}$(L1)& 0.464				& 0.250				& 0.241			\\
 $\mathrm{VILoss_{MSE}}$(L2)& 0.316				& 0.236				& 0.243			\\
 \hline
 \multicolumn{4}{c}{$\lambda=2$}													\\
 $\mathrm{VILoss_{MSE}}$(L1)& \textbf{0.300}	&\textbf{0.202}		&\textbf{0.193}	\\
 $\mathrm{VILoss_{MSE}}$(L2)& \textbf{0.277}	&\textbf{0.169}		&\textbf{0.175}	\\
 \hline
 \multicolumn{4}{c}{$\lambda=5$}													\\
 $\mathrm{VILoss_{MSE}}$(L1)& 0.345				& 0.249				& 0.204			\\
 $\mathrm{VILoss_{MSE}}$(L2)& 0.331				& 0.206				& 0.195			\\
 \hline
 \multicolumn{4}{c}{$\lambda=10$}													\\
 $\mathrm{VILoss_{MSE}}$(L1)& 0.542				& 0.278				& 0.203			\\
 $\mathrm{VILoss_{MSE}}$(L2)& 0.454				& 0.223				& 0.185			\\
 \hline
 \multicolumn{4}{c}{$\lambda=20$}													\\
 $\mathrm{VILoss_{MSE}}$(L1)& 0.340				& 0.242				& 0.198			\\
 $\mathrm{VILoss_{MSE}}$(L2)& 0.467				& 0.198				& 0.185			\\
 \hline
 \multicolumn{4}{c}{$\lambda=50$}													\\
 $\mathrm{VILoss_{MSE}}$(L1)& 0.369				& 0.323				& 0.213			\\
 $\mathrm{VILoss_{MSE}}$(L2)& 0.351				& 0.287				& 0.203			\\
 \hline
 \multicolumn{4}{c}{$\lambda=100$}													\\
 $\mathrm{VILoss_{MSE}}$(L1)& 0.463				& 0.336				& 0.193			\\
 $\mathrm{VILoss_{MSE}}$(L2)& 0.433				& 0.333				& 0.197			\\
 \hline
\end{tabular}
\label{tab:synth_1D_full}
\end{table}

% table:synth_2D_full
\begin{table}[ht]
\centering
\caption{Full test results of the error of models trained with VILoss and the base loss on Synth-2D. Poly-3 and Poly-10 models are used to show the cases when under-parameterized and over-parameterized models are trained on the data set, respectively.}
\begin{tabular}{ l l l l } 
 \hline
 	 & \multicolumn{3}{c}{test error of models}										\\
 	 						& Poly-3			& Poly-6 			& Poly-10		\\ 
 \hline
 MSE loss					& 0.378				& 0.197				& 0.207			\\
 \hline
 \multicolumn{4}{c}{$\lambda=1$ (no partitioning)}									\\
 $\mathrm{VILoss_{MSE}}$(L1)& 0.600				& 0.396				& 0.386			\\
 $\mathrm{VILoss_{MSE}}$(L2)& 0.676				& 0.470				& 0.452			\\
 \hline
 \multicolumn{4}{c}{$\lambda=2$}													\\
 $\mathrm{VILoss_{MSE}}$(L1)& 0.336				& 0.163				& 0.166			\\
 $\mathrm{VILoss_{MSE}}$(L2)& 0.289				& 0.136				& 0.138			\\
 \hline
 \multicolumn{4}{c}{$\lambda=5$}													\\
 $\mathrm{VILoss_{MSE}}$(L1)& \textbf{0.317}	& 0.145				& 0.159			\\
 $\mathrm{VILoss_{MSE}}$(L2)& \textbf{0.269}	& 0.101				& 0.115			\\
 \hline
 \multicolumn{4}{c}{$\lambda=10$}													\\
 $\mathrm{VILoss_{MSE}}$(L1)& 0.357				& \textbf{0.126}	& \textbf{0.142}\\
 $\mathrm{VILoss_{MSE}}$(L2)& 0.309				& \textbf{0.078}	& \textbf{0.092}\\
 \hline
 \multicolumn{4}{c}{$\lambda=20$}													\\
 $\mathrm{VILoss_{MSE}}$(L1)& 0.347				& 0.215				& 0.198			\\
 $\mathrm{VILoss_{MSE}}$(L2)& 0.351				& 0.178				& 0.200			\\
 \hline
 \multicolumn{4}{c}{$\lambda=50$}													\\
 $\mathrm{VILoss_{MSE}}$(L1)& 0.319				& 0.153				& 0.319			\\
 $\mathrm{VILoss_{MSE}}$(L2)& 0.296				& 0.127				& 0.331			\\
 \hline
 \multicolumn{4}{c}{$\lambda=100$}													\\
 $\mathrm{VILoss_{MSE}}$(L1)& 0.559				& 0.919				& 1.950			\\
 $\mathrm{VILoss_{MSE}}$(L2)& 0.658				& 0.937				& 1.968			\\
 \hline
\end{tabular}
\label{tab:synth_2D_full}
\end{table} 

%
%\textbf{\emph{Do not put content after the references.}}
%%
%Put anything that you might normally include after the references in a separate
%supplementary file.
%
%We recommend that you build supplementary material in a separate document.
%If you must create one PDF and cut it up, please be careful to use a tool that
%doesn't alter the margins, and that doesn't aggressively rewrite the PDF file.
%pdftk usually works fine. 
%
%\textbf{Please do not use Apple's preview to cut off supplementary material.} In
%previous years it has altered margins, and created headaches at the camera-ready
%stage. 
%%%%%%%%%%%%%%%%%%%%%%%%%%%%%%%%%%%%%%%%%%%%%%%%%%%%%%%%%%%%%%%%%%%%%%%%%%%%%%%%
%%%%%%%%%%%%%%%%%%%%%%%%%%%%%%%%%%%%%%%%%%%%%%%%%%%%%%%%%%%%%%%%%%%%%%%%%%%%%%%%

\end{document}